\documentclass{article}

\usepackage{PRIMEarxiv}

\usepackage[utf8]{inputenc} 
\usepackage[T1]{fontenc}    
\usepackage{hyperref}       
\usepackage{url}            
\usepackage{booktabs}       
\usepackage{amsfonts}       
\usepackage{nicefrac}       
\usepackage{microtype}      
\usepackage{lipsum}
\usepackage{fancyhdr}       
\usepackage{graphicx}       
\graphicspath{{media/}}     

\title{Combining multitemporal optical and SAR data for LAI imputation with BiLSTM network
}

\author{W. Zhao, F. Yin, H. Ma, Q. Wu, J. Gomez-Dans, P. Lewis}

\author{
  W. Zhao, F. Yin, H. Ma, Q. Wu, J. Gomez-Dans, P. Lewis \\
  Dept.\ of Geography \\
  University College London \\
  Gower Street, London, WC1E 6BT UK\\
  \texttt{\{weiying.zhao, feng.yin.15, hongyuan.ma, qingling.wu, j.gomez-dans, p.lewis\}@ucl.ac.uk} \\
}

\begin{document}
\maketitle

\begin{abstract}
Leaf area index (LAI) is an important biophysical parameter and plays a significant role in winter wheat yield prediction. Persistent clouds dramatically affect the acquisition of crop conditions with Sentinel-2 remote sensing images, which may cause unreliable yield predictions. Synthetic Aperture Radar (SAR) can provide all-weather imagery and the ratio between cross- and co-polarized channels (C-band) has a high correlation with time series LAI over winter wheat areas. Here, the time series Sentinel-1 VH/VV is evaluated for imputing LAI, so as to increase its spatial-temporal density. We propose using a bidirectional LSTM (BiLSTM) network to impute the time series LAI. The half mean squared error of the predicted time series for each time step is used as the loss function. During the training, the mean loss over the observations in the mini-batch is calculated. Two test regions are selected, one is in the south of German, and another is in the North China Plain. Only the Sentinel-1 VH/VV and Sentinel-2 generated LAI which is acquired over the growing season are used during the training. Sentinel-2 generated LAI, which has been calibrated by in-situ measurements, is treated as the true value during the training. Plenty of experimental results show that the LAI imputation results provided by the BiLSTM method are much better than traditional regression methods, such as exponential function and polynomial function. It can capture the nonlinear dynamics between multiple time series. BiLSTM is robust to impute the time series LAI in different winter wheat fields with different growing conditions. It can even provide satisfactory results when fewer Sentinel-2 images are available. Since LSTM only see the information from the past, it cannot provide as good results as BiLSTM, especially over the  senescence period. In conclusion, the results indicate that the BiLSTM network can be used to impute LAI with time-series Sentinel-1 VH/VV data and Sentinel-2 generated LAI data. This method can also be extended to solve other time series missing value imputation problems.
\end{abstract}


\section{Introduction}

Leaf area index (LAI) can characterize crop canopies with dimensionless quantity. Timely, continuous and accurate monitoring of crop LAI is critical for winter wheat yield forecasting. Optical sensors with  high spatial and spectral resolutions  have been widely  used for generating LAI. However,  the passive acquisition model is easily affected by cloudy weather in the crop growing period between seedling and flowering, which will lead to lots of missing values \cite{pipia2019fusing}. 
Currently, more and more available multisensor, multitemporal and multispectral data allow the estimation of missing values and fill the temporal monitoring information gaps. SAR sensors are ideal to fulfil this task, because of their all-time and all-weather
capabilities, with very good accuracy of the acquisition geometry and without effects
of atmospheric constituents for amplitude data.  Thanks to the Copernicus programme, more and more Sentinel-1 and Sentinel-2 data are acquired and free access. In this paper, we mainly pay attention to the integration of multitemporal Sentinel-1 Ground Range Detected (GRD) VH and VV ratio  (VH/VV) and limited available Sentinel-2 generated LAI to impute more LAI values.

There are three popularly used methods dealing with missing values: regression  \cite{veloso2017understanding}, interpolation and matrix completion \cite{yoon2018estimating}.
 Interpolation methods can provide a smooth result, but it needs the points which can describe the whole temporal features.
For time series Sentinel-2 generated LAI and time series Sentinel-1, they have the capacity of describing the whole growth of winter wheat, with fast increasing values in the spring or rainy season  and a decrease in the senescence period \cite{veloso2017understanding}.
  Multiple papers have proved the temporal correlation between time series vegetation descriptors (LAI and NDVI) and multitemporal Sentinel-1 data \cite{pipia2019fusing, veloso2017understanding, kim2011radar, stendardi2019exploiting}.  

Even though the winter wheat condition may vary in different fields,  multisensor time series have a high feature correlation in the same geolocation.
With the temporal correlated time series, we transfer the LAI imputation problem to a multivariate time series regression task. We apply the BiLSTM network to fulfil this task, it can take into account both the forward and backward information. 

The following section will briefly introduce BiLSTM theory. The preprocessing of multisensor time series data will be introduced in section 3. In section 4, we will evaluate the spatial and temporal imputation performance of the trained BiLSTM network and compare it with the other 3 methods. Then, the conclusion is drawn in the final section.

\section{Methodology}

BiLSTM is a popularly used non-parametric modelling  technique. In the following, we briefly introduce the general BiLSTM formulations and their application to solving the time series LAI imputation problem.

\begin{figure}
   \centering
\begin{tabular}{c}
\includegraphics[width=10cm]{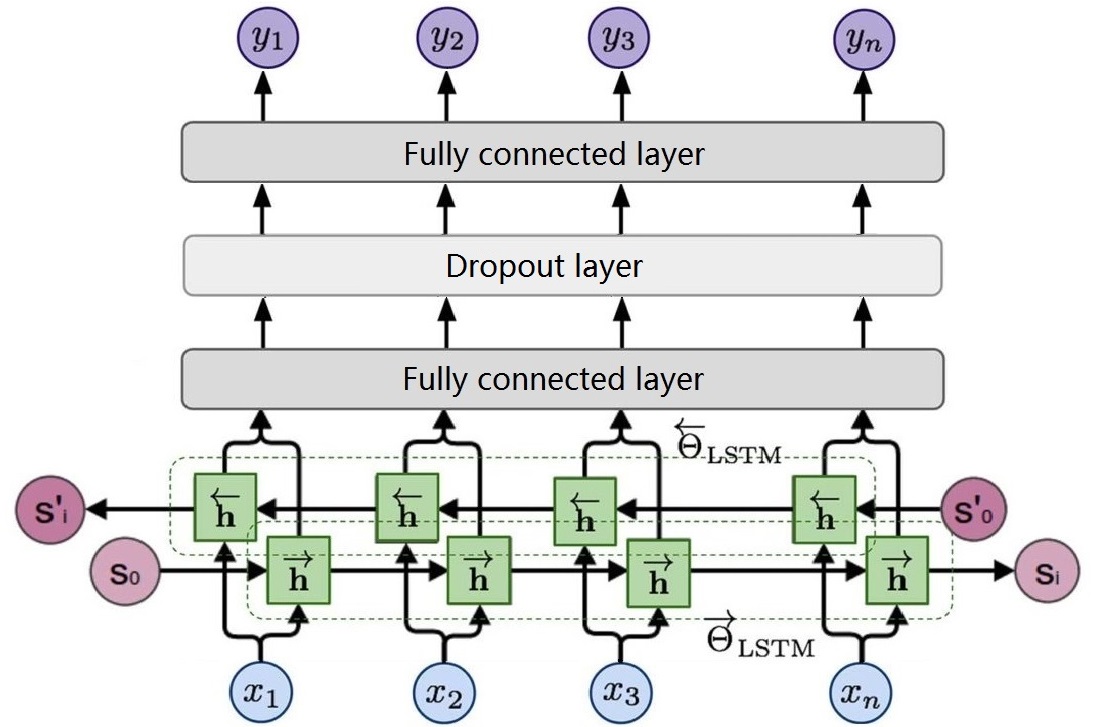} \\
\end{tabular}
  \caption{Bidirectional LSTM architecture.  $S$ represents the network state. (The architecture is modified based on Christopher Olah' blog$^1$) }
  \label{fig:BiLSTMarchitech} 
\end{figure}

The architecture of BiLSTM network  \footnote{Neural Networks, Types, and Functional Programming: http://colah.github.io/posts/2015-09-NN-Types-FP/} is shown as Fig.\ref{fig:BiLSTMarchitech}. A BiLSTM layer learns bidirectional long-term dependencies between time steps of time series. These dependencies can be useful when learning the complete time series at each time step. 
Relative insensitivity to gap length is an advantage of LSTM over RNNs and other sequence learning methods in numerous applications.

The large variance of the training data values may generate an unstable model, which will cause a large generalization error. Thus, given a time series vector $X$, we normalize the training predictors to have zero mean and unit variance. 

\begin{equation}
    x = \frac{X - \mathbf{E}(X)}{{Var(X)}}
\end{equation}

For each element in the input sequence, each layer computes the following
function:
\begin{equation}
\begin{array}{ll}
        i_t = \sigma(W_{ii} x_t + b_{ii} + W_{hi} h_{(t-1)} + b_{hi}) \\
        f_t = \sigma(W_{if} x_t + b_{if} + W_{hf} h_{(t-1)} + b_{hf}) \\
        g_t = \tanh(W_{ig} x_t + b_{ig} + W_{hg} h_{(t-1)} + b_{hg}) \\
        o_t = \sigma(W_{io} x_t + b_{io} + W_{ho} h_{(t-1)} + b_{ho}) \\
        c_t = f_t \circ c_{(t-1)} + i_t \circ g_t \\
        h_t = o_t \circ \tanh(c_t) \\
    \end{array}
\end{equation}
where $i_t$, $f_t$, $g_t$,
$o_t$ are the input gate, forget gate, cell gate, and output gate, respectively. $c_t$ is the cell
state at time $t$, $h_t$ is the hidden state at time $t$, $x_t$ is the input at time $t$,  $\sigma$ is the sigmoid function, and $\circ$ represents the Hadamard product.

Then, the network follows by a fully connected layer $ \tilde{y}_t = \sigma (Wh_t + b)$, a dropout layer to avoid the overfitting problem and a fully connected layer with the same length as the input features, so as to capture all the information of the entire sequence at each prediction.

The half-mean-squared-error of the predicted responses for each time step
is used as the imputation error:
\begin{equation}
    \mathcal{L} = \frac{1}{2M}\sum_{i=1}^M\sum_{j=1}^R (\hat{y}_{ij}-y_{ij})^2
\end{equation}
where $M$ is the sequence length and $R$ is the amount of responses.  This process acts as the regression layer and it will compute the loss over the training data in the mini-batch. Finally, we can obtain the imputed $\hat{y}$ through minimizing $\mathcal{L}(\hat{y}, y)$. 
 The architecture of LSTM only contains the information flow from the past to the forward (from $S_0$ to $S_i$ in Fig.\ref{fig:BiLSTMarchitech}).

\begin{table}
\centering
 \caption{Part of multisensor time series with missing values and imputation results. Both the multisensor time series contain the missing values which are indexed by 0.  All the valid Sentinel-2 LAI time series are larger than 0, while all the valid Sentinel-1 VH/VV values are smaller than 0. All of these values have the same geolocation. The valid values represent the growing condition of winter wheat.}
{
\begin{tabular}{llllllllllll}
\hline
\multicolumn{1}{|l|}{Imputation results} &\multicolumn{1}{|l|}{0.32} & \multicolumn{1}{l|}{0.36} & \multicolumn{1}{l|}{0.43} & \multicolumn{1}{l|}{0.53} & \multicolumn{1}{l|}{0.76} & \multicolumn{1}{l|}{0.96} & \multicolumn{1}{l|}{ 1.38} & \multicolumn{1}{l|}{1.8} & \multicolumn{1}{l|}{2.16} & \multicolumn{1}{l|}{ 2.57} & \multicolumn{1}{l|}{ 2.73} \\ \hline
\multicolumn{1}{|l|}{Sentinel-2 LAI} &\multicolumn{1}{|l|}{0.32} & \multicolumn{1}{l|}{0} & \multicolumn{1}{l|}{0.43} & \multicolumn{1}{l|}{0.53} & \multicolumn{1}{l|}{ 0} & \multicolumn{1}{l|}{0} & \multicolumn{1}{l|}{0} & \multicolumn{1}{l|}{1.8} & \multicolumn{1}{l|}{2.16} & \multicolumn{1}{l|}{0} & \multicolumn{1}{l|}{0} \\ \hline
\multicolumn{1}{|l|}{Sentinel-1 VH/VV} &\multicolumn{1}{|l|}{0} & \multicolumn{1}{l|}{-8.28} & \multicolumn{1}{l|}{0} & \multicolumn{1}{l|}{-7.85} & \multicolumn{1}{l|}{-7.56} & \multicolumn{1}{l|}{-7.21} & \multicolumn{1}{l|}{-6.84} & \multicolumn{1}{l|}{0} & \multicolumn{1}{l|}{-6.12} & \multicolumn{1}{l|}{-5.83} & \multicolumn{1}{l|}{-5.65} \\ \hline
\end{tabular}}
\label{tab:MultiSensorTimeSeries} 
\end{table}

During the application of the BiLSTM method, only the temporal points which have at least one acquisition value are used  (Tab.\ref{tab:MultiSensorTimeSeries}). 
In the spatial domain, the length of the time series may be different. After standardization, the missing values will be treated as one kind of information. We regard missing values as variables of the BiLSTM graph, which are involved
in the backpropagation process \cite{cao2018brits}.

\section{Study area and time series preprocessing} 

In this section, we introduce the preparation of the multisensor time series. Two areas are selected to train and evaluate the methods. One is located in the north of Munich, Germany.  The other one is located in Hengshui area, Hebei Province, China. All the Sentinel images are prepared using Google Earth Engine. The preprocessing of Sentinel-1 and Sentinel-2 images is the same as that used in the Sentinel toolbox. The future processing of the data will be introduced as follows.

\subsection{Preprocessing of Sentinel-2 LAI time series}

 Sentinel-2 is systematically acquired at high spatial resolution (10 m) with 5 revisiting days (with 2 satellites) in the same observation angle. Thanks to  Copernicus Programme, it is open access. 
After atmospheric correction of Sentinel-2 TOA reflectance, we can retrieve LAI using inverse emulator \footnote{SIAC on GEE: https://github.com/MarcYin/SIAC\_GEE}, as shown in \cite{YinSIAC19}.
All the time series LAI over the test areas are generated by Sentinel-2 images.

Only the areas in the Sentinel-2 images which are less affected by the clouds are selected. 10 Sentinel-2 images are selected which were acquired  over the Hengshui area between March 2017 and June 2017.  12 Sentinel-2 images are used which were acquired  between March 2017 and August 2017 over the North of Munich.

\subsection{Preprocessing of Sentinel-1 time series}

The Sentinel-1 GRD images which are obtained through interferometric wide  swath mode over the same area during a similar acquisition period as Sentinel-2 images are collected. 
To reduce the speckle effect, we use the multitemporal denoised images \cite{quegan2001filtering} to compute Sentinel-1 VH and VV ratio.
Specifically, given a time series of $M$ dB values $[I_1(s), I_2(s), I_3(s), \cdot\cdot\cdot,I_M(s)]$ indexed by time $t$, the denoised value at location $s$ can be calculated through:

\begin{equation}
\hat{J}_t(s)=\frac{\mu^{{ML}}_t(s)}{M} \sum_{t'=1}^{M} \frac{\emph{I}_{t'}(s)}{\mu^{{ML}}_{t'}(s)}
\end{equation}
where $\emph{I}_t(s)$ is the calibrated dB value, $J_t(s)$  the noise-free value, $\mu^{{ML}}_t(s)$  is the initial estimation value using multi looking (with a window size of 14$\times$14), $s$ is the position in one image and $t$ is the serial number of $M$ images. Then, the variance of the denoised value can be computed  with:
\begin{equation}
    {Var}_t(s) = \hat{J}^{{ML}}_t(s)\frac{M+N-1}{MNL}
\end{equation}
Finally, we can compute the dual polarization ratio with is computed through $ {\hat{J}^{{VH}}_t(s)}/{\hat{J}^{{VV}}_t(s)}$ \cite{veloso2017understanding}.

To ensure all the time series SAR data $I_t(s)$ capture the same signal from the object, we only use the images which are acquired through a similar incidence angle. 214 images (orbit 117)  are used for multitemporal SAR image denoising over the Munich test area, while more than 220 images acquired through ascending orbits over the Hengshui test area are used.   
All the SAR images are well-registered with the Sentinel-2 images which are obtained at the same place.

\section{Experimental results and discussion}
In this section, we demonstrate the spatial and temporal imputation performance of the BiLSTM network by comparing  it with some popularly used methods. 

In the test areas, most of the Sentinel-2 LAI and Sentinel-1 VH/VV values are acquired at different times.  To keep the spatial resolution of LAI, polynomial and exponential regression parameters are computed once for each time series. During the imputation, Sentinel-1 VH/VV time series are interpolated according to Sentinel-2 acquisition time. Since non-LSTM methods cannot handle arbitrary lengths of the time series, we only compare them with the test of real-time series. 

LSTM and BiLSTM have the same parameters before training. 60 hidden units are used, followed by a fully connected layer of size 50, a dropout layer with a 0.5 probability of dropping out input elements and a fully connected layer with the same length as the input features. The depth will be good for well-leaning longer-term dependencies,  that is partly the reason people stack LSTMs in sequence-to-sequence models.
The Sentinel acquisition frequency is very high over Germany test areas, they can show all the temporal dynamic features of winter wheat. We will use them to train the networks. During the whole growing period, the length of the time series may be different in different places, either caused by the weather or the sensor's acquisition settings. To ensure the trained network is robust to different situations, the training time series has been randomly sampled with the shortest length equal to 8 and the longest length equal to 69.  Only the pixels which correspond to winter wheat are used.

\subsection{Temporal imputation}

\begin{figure}
\setlength{\abovecaptionskip}{-10pt} 
\setlength{\belowcaptionskip}{-10pt}
\centering
\includegraphics[width=16cm]{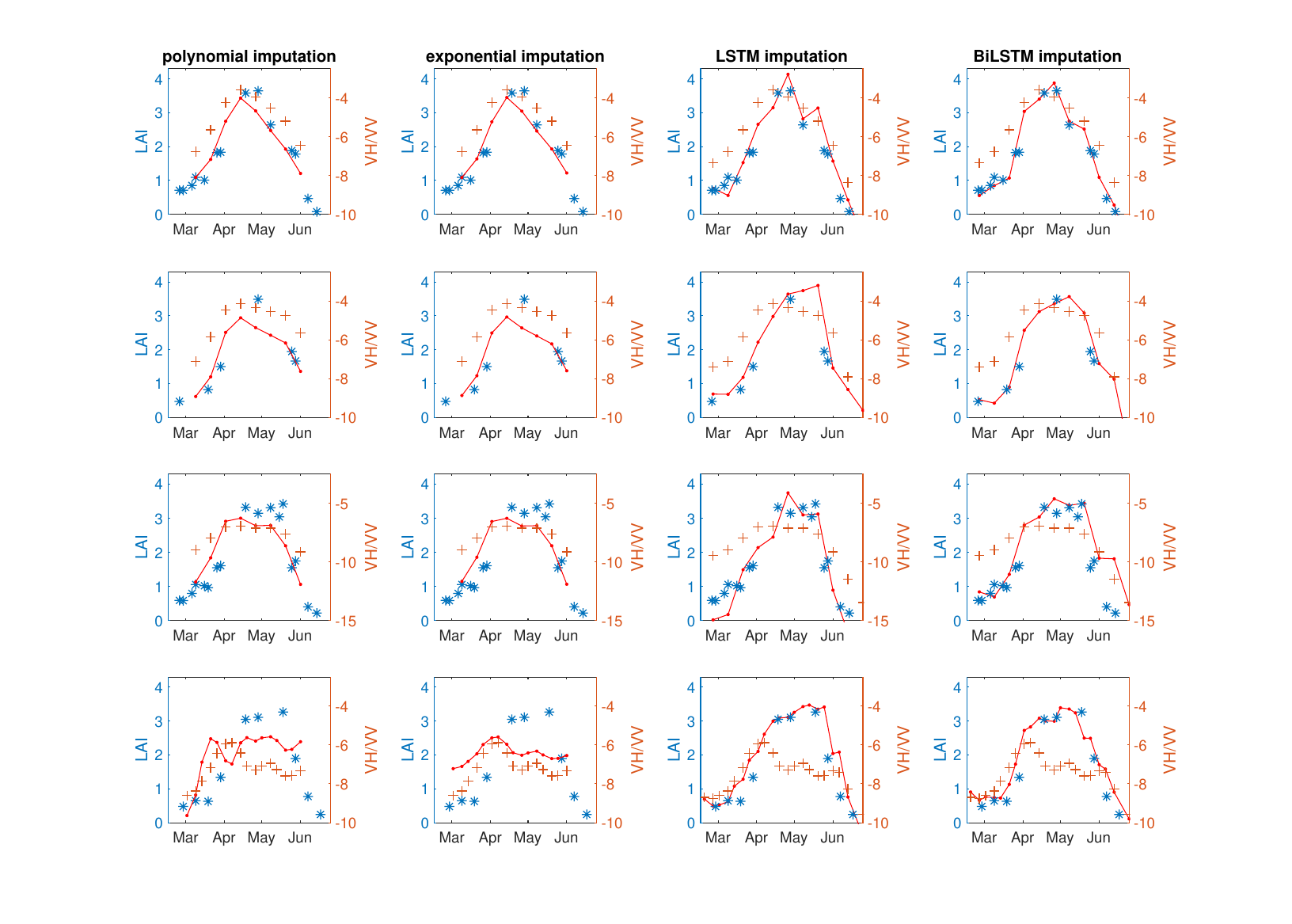} \\
\caption{Different time series LAI imputation results in comparison. From left to right: polynomial regression, exponential regression, LSTM, and BiLSTM. Imputed results are shown using the red line.  Four different places (four rows) in the North China Plain are used. The imputation results and index using the red line. Sentinel-1 VH/VV time series are smoothed using a Gaussian kernel with a window size equal to 5.
All the SAR data in the first three rows are acquired through the same orbit, while the SAR data in the fourth row are acquired through the ascending track with an orbit number equal to 40 or 142.}
  \label{fig:TimeSeriesImputationComp} 
\end{figure}

4 different time series (Fig.\ref{fig:TimeSeriesImputationComp}) are selected to show the imputation difference provided by the different methods. There are some gaps in the LAI time series between April and May, which correspond to the rainy season. Sentinel-2 images cannot capture the crop growing conditions because of the cloudy weather. However, the LAI values acquired over this period are very important for crop yield prediction. 

For the fourth test area (fourth row of Fig.\ref{fig:TimeSeriesImputationComp}), the saturation of SAR backscattering values may be caused by the similar side-looking angle and sowing direction. This similarity may cause SAR sensors to capture low-volume backscattering values or by the large multi-looking window size during the multitemporal single-channel denoising. This phenomenon popularly exists in the North China Plain, especially when all the ascending orbits acquiring data are used for the multitemporal denoising.

When the multisensor time series have similar temporal features, polynomial and exponential imputation methods can provide satisfying results. These two methods are much simpler and faster than LSTM and BiLSTM. However, when the two features have a complex relationship, they can not provide good results, as shown in the fourth row of Fig.\ref{fig:TimeSeriesImputationComp}. When the training samples contain too much noise.
The multisensor time series have a nonlinear relationship which is too hard to describe using a low-degree regression model.

\begin{table}
\centering
\caption{LAI imputation performance using different methods. 20 points are used for the evaluation. The 20 points are distributed in the  Hengshui area with two times field measurements. One is in March, another is in May. }
\label{Tab:RMSEresults}
\begin{tabular}{ccc}
\toprule
\textbf{Methods}& RMSE (March)  & RMSE (May) \\
\midrule
 BiLSTM& 0.68&	1.7855 \\
 LSTM &0.5912&	2.2886 \\
 Polynomial &0.9339&	2.4088 \\
 Exponential &0.9361&	2.624 \\
\bottomrule
\end{tabular}
\end{table}

Generally, people always compute the regression coefficients between one feature and another, and find the optimized parameter based on the least-squares estimation. This will lead to the inverse results smaller than the referenced target feature in the peak part. 

LSTM and BiLSTM can take into account the two features and generate better imputation results. As we can see from Fig.\ref{fig:TimeSeriesImputationComp}, all the imputation results provided by LSTM and BiLSTM are better than the former two methods. Even when Sentinel-1 signals have complex changes, they still provide good results.  During the imputation, they can take the limited number of LAI points as the weight to adjust the imputed time series.

\subsection{Spatial performance of the imputation}
All the imputation is performed in the temporal domain for each geolocated point. In this section, we will analyze the imputation spatial performance by comparing the results with Sentinel-2 provided LAI image and some in-situ measurements.
During the imputation, the parameters of polynomial and exponential regression methods are separately computed for each time series. 

As observed in Tab.\ref{Tab:RMSEresults}, both BiLSTM and LSTM obtain smaller root-mean-square error (RMSE) results. With the reference of Fig \ref{fig:TimeSeriesImputationComp}, the big RMSE results provided by LSTM may be caused by the larger imputed LAI values in May. The big RMSE values provided by polynomial and exponential regression methods are caused by the lower imputation results.

\begin{figure}
   \centering
\begin{tabular}{ccc}
\includegraphics[width=5cm]{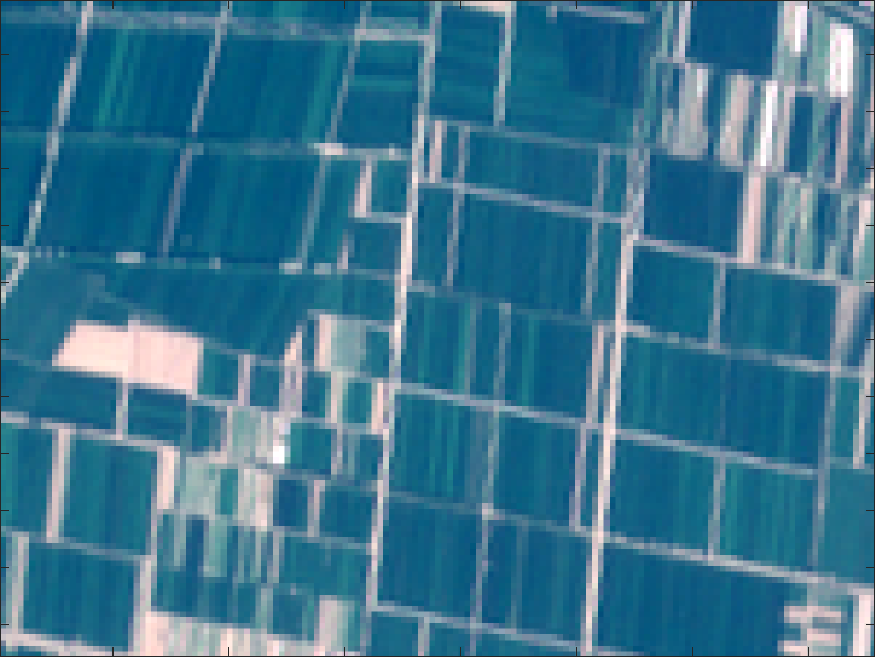} &
\includegraphics[width=5cm]{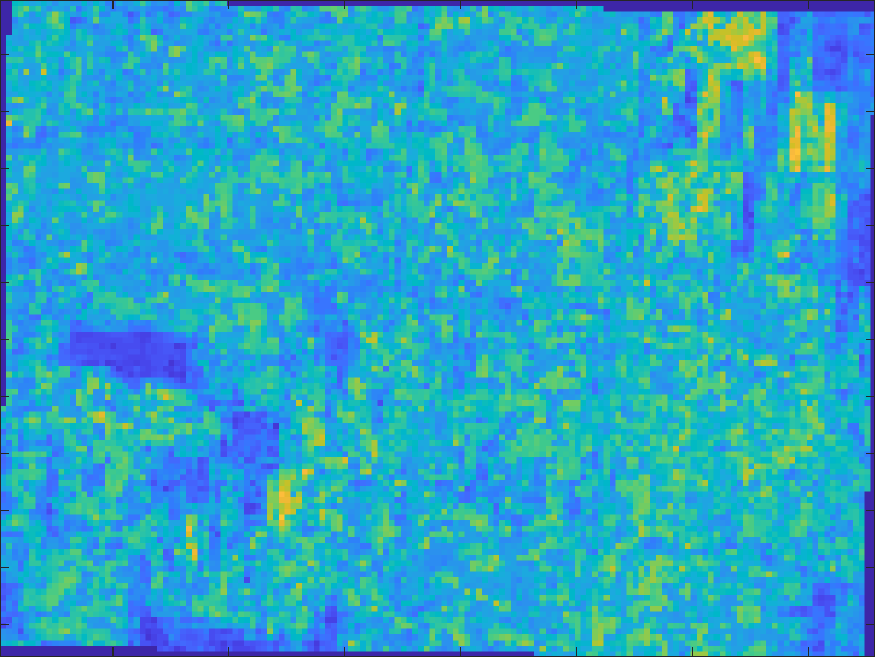} &
\includegraphics[width=5cm]{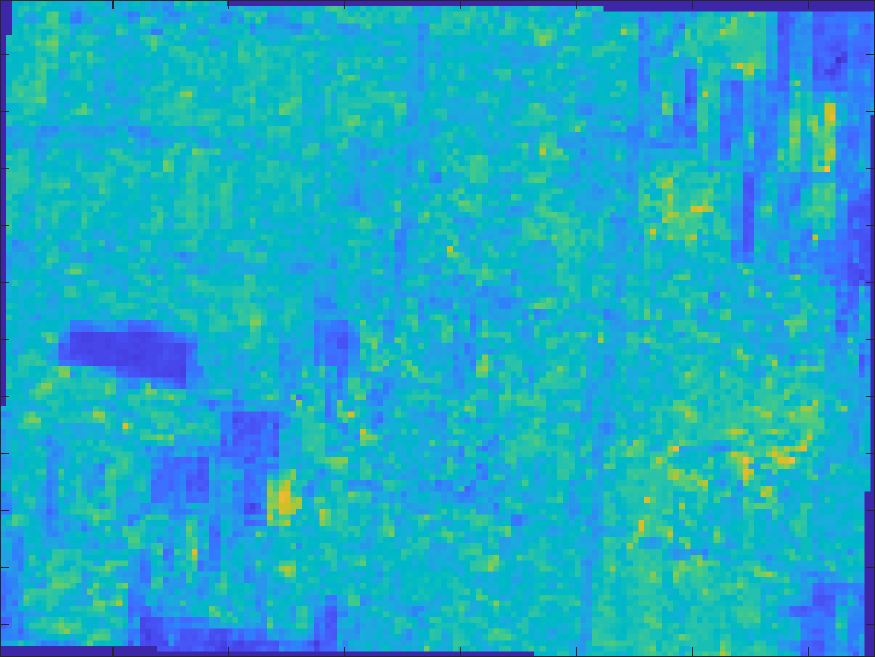}\\
(1) Sentinel-2 RGB image & (2) Polynomial regression& (3) Exponential regression\\
\includegraphics[width=5cm]{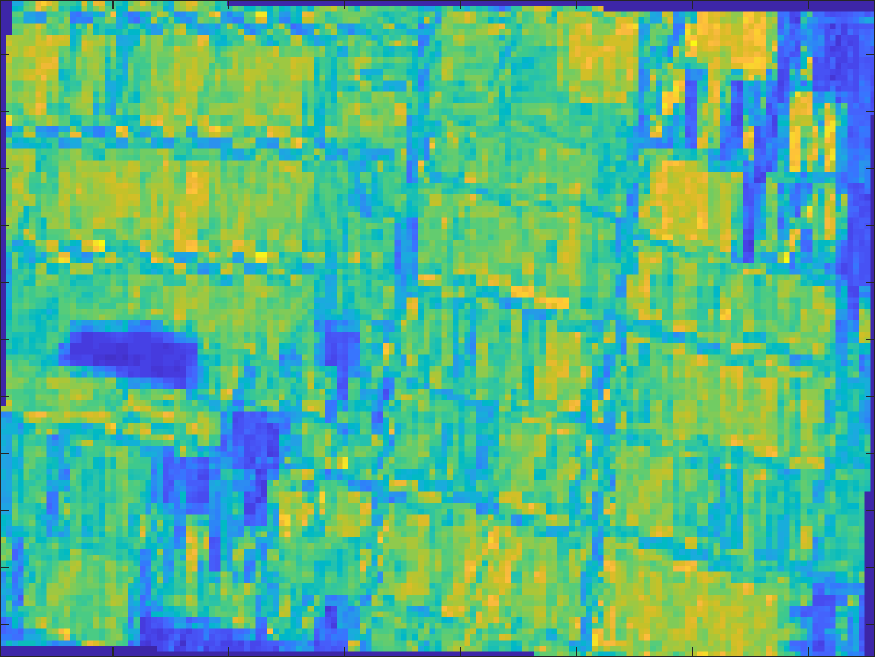} &
\includegraphics[width=5cm]{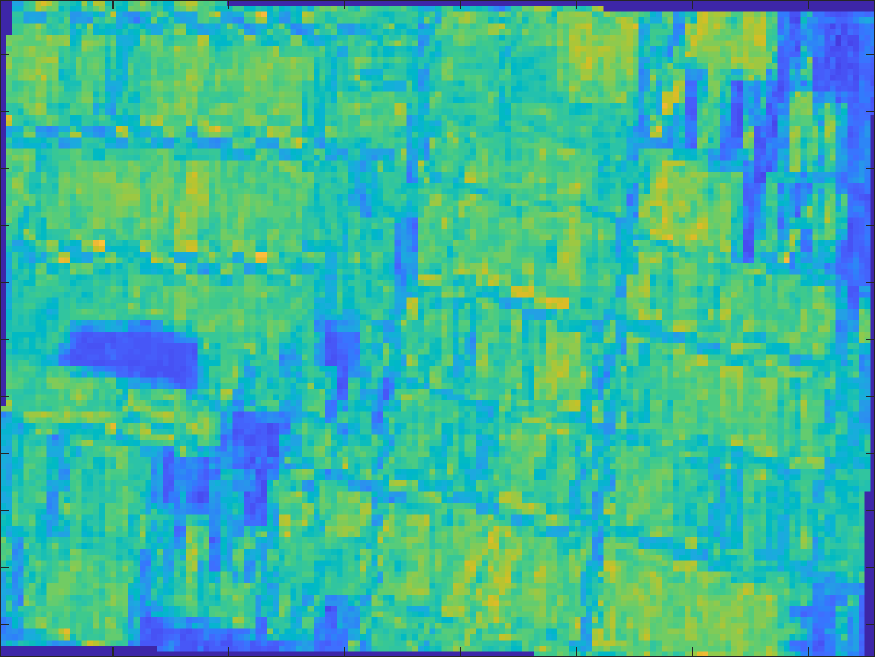} &
\includegraphics[width=5cm]{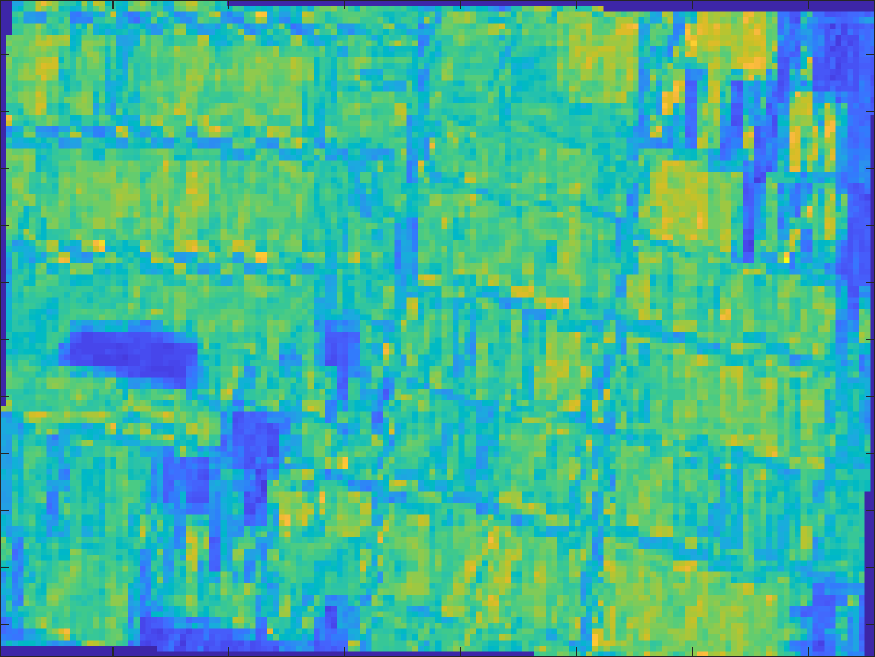} \\
(4) Sentinel-2 LAI & (5) LSTM network& (6) BiLSTM network\\
\multicolumn{3}{c}{0 \includegraphics[width=6cm]{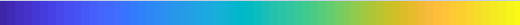} 6}\\
\multicolumn{3}{c}{\includegraphics[width=15cm]{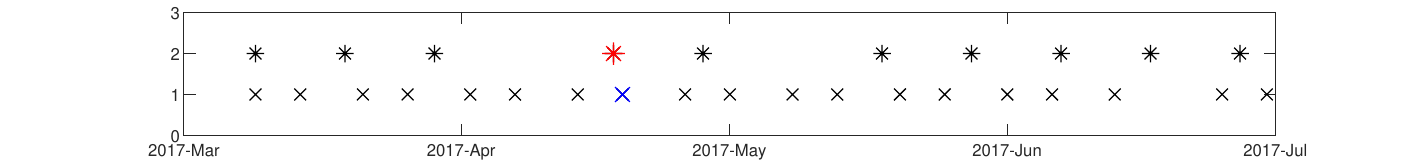}}\\
\end{tabular}
  \caption{Different LAI imputation results in comparison. The acquisition time of the Sentinel time series is shown at the bottom of the figure, with * representing Sentinel-2 acquisition time and $\times$ representing Sentinel-1 acquisition time.
  Sentinel-2 image (1, 4) is acquired on 18/04/2017 (red *) over North China Plain, and Sentinel-1 image is acquired on 19/04/2017 (blue $\times$). LAI values range from 0 to 6. }
  \label{fig:LAIimputationResults} 
\end{figure}

From Fig.\ref{fig:LAIimputationResults} we can see that
LSTM and BiLSTM can provide much better imputation results in the spatial domain than polynomial and exponential regression methods. Their imputation results have better spatial features, with smooth values in the same field and clear line features in the boundary areas. Compared with LSTM-provided results (Fig.\ref{fig:LAIimputationResults}(5)), BiLSTM can also provide better results in some narrow fields (Northeast of the test area in Fig.\ref{fig:LAIimputationResults}). The spatial distribution of the results provided by BiLSTM is much more similar to the real LAI.

\section{Conclusion}
In this paper, we successfully applied the BiLSTM network to impute LAI with multivariate time series data. Plenty of experiments show the effectiveness of its imputation over the spatial and temporal domains.
It is obvious that both BiLSTM and LSTM imputation performance outperforms traditional regression methods. It can well impute the missing values in the time series, especially during the fast-growing season of winter wheat. Even when the temporal Sentinel-1 VH/VV values have a complex change, the BiLSTM network still could provide effective imputation results. In the future, we will take the spatial correlation information into account for missing value imputation. The spatiotemporal multisensor LAI imputation process has been separated into several steps, we will find ways to merge them.

\bibliographystyle{unsrt}  
\bibliography{references}

\end{document}